\documentclass{article}
\usepackage[utf8]{inputenc} 
\DeclareUnicodeCharacter{202F}{\ } 

\usepackage{spconf,amsmath,graphicx, amssymb}
\usepackage{hyperref}

\usepackage{enumitem}
\setlist{nosep, leftmargin=14pt}

\usepackage{mwe} 

\usepackage{bm}

\title{Variance-Penalized MC-Dropout as a Learned Smoothing Prior for Brain Tumour Segmentation}
\name{Satyaki Roy Chowdhury and Golrokh Mirzaei}
\address{Department of Computer Science and Engineering, \\
The Ohio State University, Columbus, USA}
\vspace{3pt}
\begin{document}
%
\maketitle

Brain tumor segmentation is essential for diagnosis and treatment planning, yet many deep learning approaches, particularly convolutional neural networks, such as UNet and its variations produce noisy boundaries in regions of tumor infiltration. We introduce UAMSA-UNet, an Uncertainty-Aware Multi-Scale Attention-based Bayesian U-Net that leverages Monte Carlo Dropout to impose a data-driven smoothing prior while fusing multi-scale features and attention maps to capture both fine details and global context. The smoothing-regularized loss augments binary cross-entropy with a variance penalty across stochastic forward passes, suppressing spurious fluctuations and yielding spatially coherent masks. On BraTS2023, UAMSA-UNet improves Dice by up to 3.3\% and mean IoU by up to 2.7\% over U-Net; on BraTS2024, it delivers up to 4.5\% Dice and 4.0\% IoU gains over the best baseline. It also reduces FLOPs by 42.5\% relative to U-Net++ while maintaining higher accuracy. By coupling multi-scale attention with a learned smoothing prior, UAMSA-UNet achieves higher-quality, more efficient segmentations and provides a flexible foundation for future integration with transformer modules.


\section{Introduction}
Brain tumors, one of the most prevalent and lethal types of cancer, develop from
glial cells and present a significant risk to human health. Magnetic Resonance
Imaging (MRI) is widely used for detecting these tumors because of its excellent soft tissue contrast and non-invasive nature. With advances in computer technology and deep learning techniques, a variety of automated methods for brain tumor segmentation have been developed \cite{mirzaei2018segmentation}. U-Net has been widely adopted for medical image segmentation due to its ability to capture hierarchical spatial features, leading to the development of several enhanced variations aimed at improving performance and adaptability \cite{chowdhury2025hybridunet}. The nnU-Net framework \cite{isensee2021nnunet} extends U-Net by incorporating data augmentation, region-specific training, and advanced post-processing, yielding improved segmentation performance. Recognizing the Transformer's powerful capability in capturing long-range dependencies \cite{vaswani2017attention}, several studies have applied Transformer models to medical image segmentation \cite{she2023eoformer}. Notably, the TransBTS model \cite{wang2021transbts} combines Transformer modules with 3D CNNs for effective spatial feature extraction.

While deep learning-based approaches have achieved impressive accuracy and detection rates in MRI-based brain tumor segmentation, they often struggle to precisely delinate tumor boundries, which remains a key limitation in segmenting regions of interest accurately. Predicting uncertainty estimation can enhance model reliability by quantifying confidence levels at the per-pixel level. In particular, epistemic uncertainty highlights regions where the model lacks knowledge, indicating areas that require further refinement \cite{kendall2017uncertainties}. Uncertainty estimation in segmentation has been explored in medical imaging \cite{carannante2021trustworthy, tang2022unified}. However, most research has primarily focused on larger anatomical structures, such as the liver, kidney, and pancreas, while comparatively less attention has been given to segmenting smaller, more heterogeneous tumors. To address this, Zhou et al. \cite{zhou2023uncertainty} proposed a multi-modal feature fusion strategy combined with Bayesian deep learning to capture segmentation uncertainties in brain tumor segmentation. The generated uncertainty maps were fed into a segmentation refinement network to enhance model robustness. Similarly, uncertainty-aware training has been applied in a semi-supervised setting to segment the left atrium \cite{yu2019uncertainty}, demonstrating its potential for improving segmentation reliability in challenging medical imaging tasks. While prior works \cite{zhou2023uncertainty, yu2019uncertainty} have employed attention mechanisms to enhance segmentation performance, they typically operate at a single scale, limiting their ability to capture both fine-grained local details and broader global context. Additionally, existing methods do not fully exploit the complementary benefits of multi-scale feature representations. To address these limitations, we propose a novel multi-scale attention module that effectively integrates local and global contextual information, providing a more comprehensive feature representation. Although we motivate the variance penalty as an uncertainty-aware term, in practice it behaves as a learned smoothing prior on pixelwise predictions. By discouraging excessive fluctuation across MC-Dropout samples, the model learns to produce spatially coherent masks, akin to Total Variation regularization but data-driven through the stochastic forward passes.

The main contributions of this work are twofold. First, We propose a Bayesian segmentation framework that leverages MC dropout to learn a data-driven smoothing prior and fuse multi-scale features for both local detail and global context. Second, we introduce a smoothing-regularized loss that augments binary cross-entropy with a variance-based penalty across stochastic forward passes to suppress spurious fluctuations in low-confidence regions, producing spatially coherent masks with improved precision and consistency.

\section{Methodology}
The overall architecture of UAMSA-UNet is illustrated in Fig. 1. It extends the U-Net encoder–decoder backbone with dropout-enabled ConvBlocks for approximate Bayesian inference and a Multi-Scale Attention Module (MSAM) to enhance representation across different spatial scales. Stochastic forward passes generate uncertainty-aware predictions, while a smoothing-regularized loss encourages spatial consistency. Together, these components form an integrated framework for uncertainty-aware, multi-scale tumor segmentation.

\begin{figure*}[h!]
    \centering
    \includegraphics[height=10.5cm, width=15.5cm]{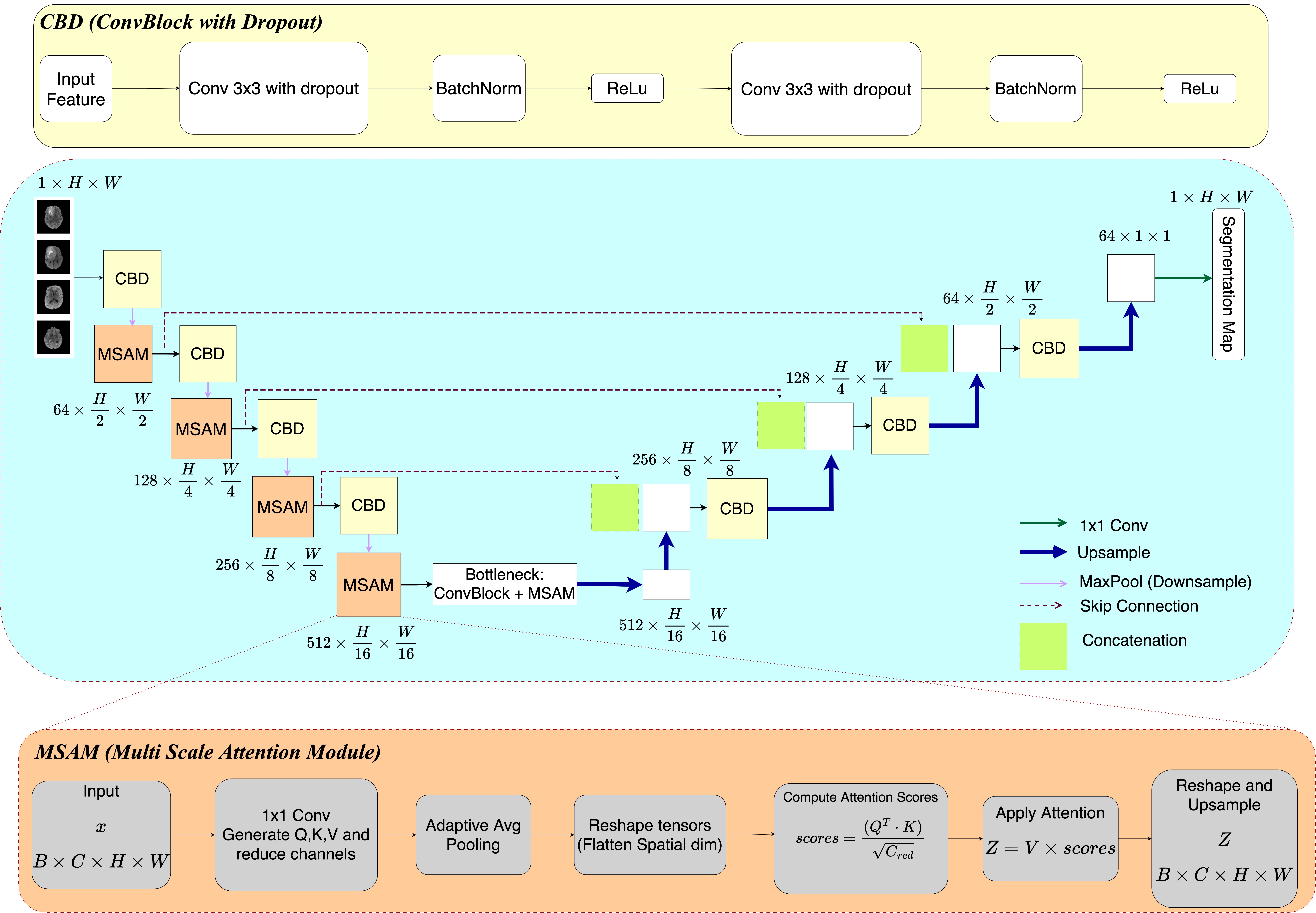}
    \caption{Overview of the proposed Bayesian U-Net with Multi-Scale Attention Module (MSAM). The model follows a UNet style encoder-decoder architecture, with convolutional blocks and  Monte Carlo dropout for Bayesian uncertainty estimation. The encoder integrates MSAM at each layer to enhance feature refinement by emphasizing on the most relevant regions while preserving spatial information and passing the information to the decoder through skip connections.}
    \label{fig:1}
\end{figure*}

\subsection{MULTI SCALE ATTENTION MODULE}
Let $\bm{x} \in \mathbb{R}^{B \times C \times H \times W}$ denote the input feature map. 
To obtain the query, key, and value representations, three parallel $1{\times}1$ convolutions are applied:
\[
\bm{Q}, \bm{K} \in \mathbb{R}^{B \times C_{\text{red}} \times H \times W}, \quad
\bm{V} \in \mathbb{R}^{B \times C \times H \times W}, \quad
C_{\text{red}} = \tfrac{C}{r}.
\]
where $r$ denotes the reduction factor. Subsequently, adaptive average pooling reduces the spatial resolution to $H' = H/4$ and $W' = W/4$, yielding compact representations:
\[
\bm{Q}', \bm{K}' \in \mathbb{R}^{B \times C_{\text{red}} \times H' \times W'}, \quad
\bm{V}' \in \mathbb{R}^{B \times C \times H' \times W'}.
\]
After reshaping with $N = H'W'$, we obtain:
\[
\hat{\bm{Q}}, \hat{\bm{K}} \in \mathbb{R}^{B \times C_{\text{red}} \times N}, \quad
\hat{\bm{V}} \in \mathbb{R}^{B \times C \times N}.
\]
The attention map and context-enhanced feature are then computed as:
\begin{equation}
\bm{A} = \mathrm{softmax}\!\left(\frac{\hat{\bm{Q}}^{\top}\hat{\bm{K}}}{\sqrt{C_{\text{red}}}}\right) \in \mathbb{R}^{B \times N \times N}, 
\quad
\hat{\bm{Z}} = \hat{\bm{V}}\bm{A}.
\end{equation}
The output $\hat{\bm{Z}}$ is reshaped to $\bm{Z}' \in \mathbb{R}^{B \times C \times H' \times W'}$ and upsampled to the original spatial resolution $\bm{Z} \in \mathbb{R}^{B \times C \times H \times W}$.
To jointly capture fine-scale details and long-range dependencies, we introduce a dual-scale attention mechanism:
\begin{itemize}
\item \textbf{Original-scale branch:} Computes attention directly on $\bm{x}$ to preserve high-frequency structural cues.
\item \textbf{Downsampled-scale branch:} Applies attention after $2\times$ spatial downsampling (i.e., $0.5\times$ resolution) to model broader context, then upsamples and fuses with the original-scale output.
\end{itemize}
Finally, both representations are fused through channel-wise concatenation followed by a $1{\times}1$ convolution:
\begin{equation}
\bm{Z}_{\text{fused}} = \mathrm{Conv}_{1{\times}1}\!\bigl(\mathrm{concat}(\bm{Z}_{\text{orig}}, \bm{Z}_{\text{down}})\bigr).
\end{equation}
This dual-scale attention formulation effectively harmonizes local discriminative features with global contextual awareness, yielding more robust and semantically consistent segmentation results.

\subsection{MONTE CARLO DROPOUT AS AN APPROXIMATE BAYESIAN INFERENCE}
Bayesian Neural Networks (BNNs) estimate uncertainty by inferring a posterior distribution over network weights~\cite{gal2016dropout}. Since exact inference is computationally intractable, variational inference approximates the posterior \( p(\mathbf{W}\mid\mathcal{D}) \) using a variational distribution \( q(\mathbf{W}) \), optimized via Evidence Lower Bound (ELBO):
\begin{equation}
    ELBO = \mathbb{E}_{q(\mathbf{W})} \left[ \log p(Y \mid X, \mathbf{W}) \right] - \mathrm{KL}\bigl(q(\mathbf{W}) \| p(\mathbf{W})\bigr).
\end{equation}
Gal and Ghahramani~\cite{gal2016dropout} demonstrated that dropout-based training serves as an approximation to variational inference, treating network weights as \(\mathbf{W} = \mathbf{M} \cdot \mathrm{diag}(\mathbf{z})\), where \(\mathbf{M}\) is a deterministic weight matrix and \(\mathbf{z}\) follows a Bernoulli distribution. At test time, Monte Carlo sampling via dropout approximates the predictive distribution as:

\begin{equation}
    p(y^* \mid x^*, \mathcal{D}) \approx \frac{1}{T} \sum_{t=1}^{T} p\bigl(y^* \mid x^*, \hat{\mathbf{W}}_t\bigr),
\end{equation}
where each \( \hat{\mathbf{W}}_t \) is sampled using dropout, allowing the model to estimate uncertainity. Kendall and Gal~\cite{kendall2017uncertainties} further distinguished epistemic uncertainty (due to model parameters) and aleatoric uncertainty (due to data noise), emphasizing their importance in robust decision-making. MC dropout primarily captures epistemic uncertainty providing valuable insights into model confidence, particularly in regions of ambiguous or complex tumor boundaries.

\subsection{SMOOTHING REGULARIZED LOSS}

To jointly optimize segmentation accuracy and uncertainty estimation, we adopt a smoothing regularized loss. Let \( p \in [0,1] \) be the predicted segmentation map (via sigmoid), and \( t \in \{0,1\} \) the ground truth. Using MC Dropout \cite{gal2016dropout}, we perform \( T \) stochastic forward passes during training. Denoting the \( t \)-th prediction as \( p^{(t)}(x) \), the mean prediction and pixel-wise uncertainty are:

\begin{equation}
\bar{p}(x) = \frac{1}{T} \sum_{t=1}^{T} p^{(t)}(x), \quad
U(x) = \frac{1}{T} \sum_{t=1}^{T} \left( p^{(t)}(x) - \bar{p}(x) \right)^2.
\end{equation}

The total loss is:
\begin{equation}
\mathcal{L}_{\mathrm{UA}} = \mathcal{L}_{\mathrm{BCE}}(p, t) + \lambda_u \cdot \frac{1}{N} \sum_{i=1}^{N} U_i,
\end{equation}
where, $\lambda_u \cdot \frac{1}{N} \sum_{i=1}^{N} U_i$ can be seen as a smoothing prior because it penalizes high-frequency variance across the 
$T$ MC-Dropout passes. In effect, we are learning a spatial regularizer that reduces spurious fluctuations while preserving meaningful boundaries.

\section{EXPERIMENTS AND RESULTS}
\subsection{DATASET AND IMPLEMENTATION DETAILS}
We evaluate our proposed UAMSA-UNet using two publicly available brain tumor segmentation datasets: BraTS2023 \cite{baid2021rsna} and BraTS2024 \cite{deverdier2024brats} adult glioma dataset. Both datasets are partitioned into training, testing, and validation sets in an 8:1:1 ratio. Since axial slices offer enhanced visibility of tumor boundaries and better capture tumor infiltration into surrounding brain regions, we extract axial slices from T1 contrast, T2-weighted, and FLAIR sequences for each patient. All images are resized to a uniform resolution of 240×240 pixels. Adam optimizer is used with a learning rate of 0.001, and the CosineAnnealingLR \cite{loshchilov2017sgdr} as the scheduler with a minimum learning rate of 1e-6. The models were trained for a total of 100 epochs with a batch size of 32. The other models are trained with Binary Cross Entropy Loss. For uncertainty estimation needed for the loss computation, we set T to 20; that is, 20 stochastic forward passes were performed to get the uncertainty term. To evaluate our method, we employ Mean Intersection over Union (mIoU) and Dice Similarity Coefficient as metrics. All the experiments are conducted using a NVIDIA GeFORCE RTX 3090 GPU with 26 GB RAM. After thorough experimental trials, the hyper-parameter $\lambda_u$ was set to 0.12 for T1C images and 0.191 for T2 and FLAIR images with a dropout probability of 0.2 in the network.

\begin{table*}[t]
    \caption{Comprehensive comparison of segmentation accuracy (Dice, mIoU) and hardware metrics across the datasets. TR: Training Runtime (sec), IT: Inference Time (ms), PWR: Power (W), MU: Memory Usage (GB), FLOPs: Floating Point Operations (G) for each model are measured on the BraTS 2023 dataset. The bold indicates the best value.}
    \label{tab:full_comparison}
    \centering
    \setlength{\tabcolsep}{0.45em}
    \resizebox{\textwidth}{!}{
    \begin{tabular}{|l|cc|cc|cc|cc|cc|cc||ccc|}
        \hline
        \textbf{Model} 
            & \multicolumn{6}{c|}{\textbf{BraTS2023}} 
            & \multicolumn{6}{c||}{\textbf{BraTS2024}} 
            & \textbf{PWR} & \textbf{MU} & \textbf{FLOPs} \\
        \cline{2-13}
        & \multicolumn{2}{c|}{T1C} 
        & \multicolumn{2}{c|}{T2 FLAIR} 
        & \multicolumn{2}{c|}{T2W}
        & \multicolumn{2}{c|}{T1C} 
        & \multicolumn{2}{c|}{T2 FLAIR} 
        & \multicolumn{2}{c||}{T2W}
        & (W) & (GB) & (G) \\
        \hline
        & Dice / mIoU & TR / IT 
        & Dice / mIoU & TR / IT 
        & Dice / mIoU & TR / IT 
        & Dice & mIoU 
        & Dice & mIoU 
        & Dice & mIoU 
        & & & \\
        \hline
        UNet \cite{unet_ref} 
            & 63.15 / 54.21 & 8476 / 13.6 
            & 70.46 / 68.28 & 8488 / 13.8 
            & 70.33 / 64.72 & 1560 / 13.8 
            & 54.33 & 45.72 
            & 63.57 & 53.59 
            & 65.87 & 55.35 
            & \textbf{190} & 24.83 & 90.5 \\
        Att. UNet \cite{oktay2018attention} 
            & 64.71 / 54.15 & 8136 / 14.5 
            & 75.33 / 68.79 & 7885 / 14.3 
            & 71.63 / 66.70 & 1778 / 13.0 
            & 54.46 & 45.03 
            & 66.19 & 55.89 
            & 70.77 & 59.93 
            & 200 & \textbf{24.60} & 84.18 \\
        UNet++ \cite{zhou2018unetplusplus} 
            & 65.92 / 53.14 & 3502 / 14.3 
            & 76.65 / 69.52 & 3496 / 13.6 
            & 74.24 / 66.65 & 3521 / 13.6 
            & 54.90 & 45.27 
            & 64.34 & 54.47 
            & 69.26 & 59.31 
            & 250 & 24.73 & 66.18 \\
        UAMSA-UNet (ours) 
            & \textbf{66.28 / 56.86} & 9967 / \textbf{2.1} 
            & \textbf{77.40 / 70.46} & 10004 / \textbf{2.1} 
            & \textbf{77.45 / 70.20} & 10078 / \textbf{2.2} 
            & \textbf{58.81} & \textbf{48.56} 
            & \textbf{67.38} & \textbf{57.69} 
            & \textbf{73.26} & \textbf{63.33} 
            & 300 & 25.22 & \textbf{52.02} \\
        \hline
    \end{tabular}
    }
\end{table*}

\subsection{RESULTS AND DISCUSSIONS}

\begin{figure}
    \centering
    \includegraphics[width=1\linewidth]{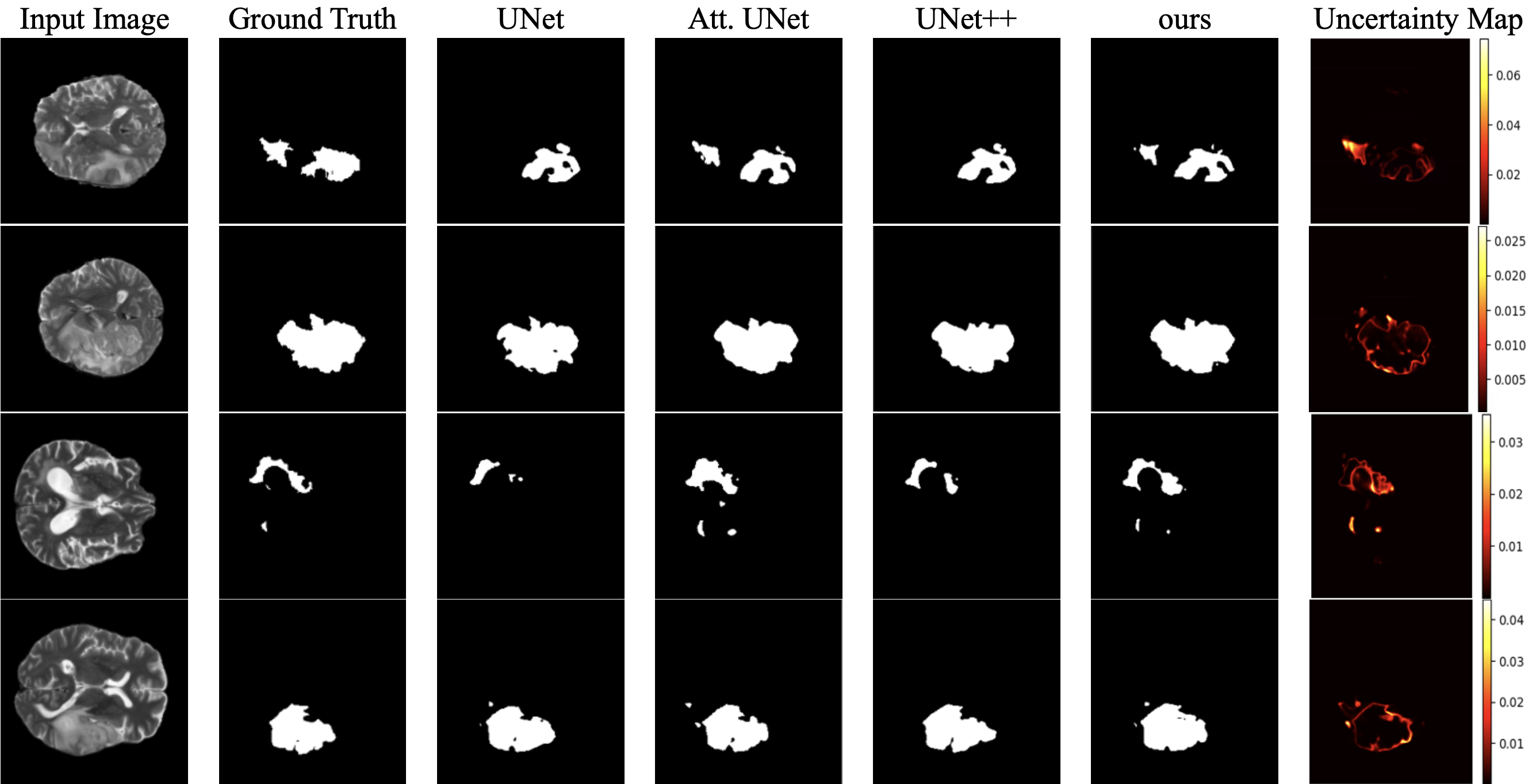}
    \caption{Qualitative comparison on BraTS2023 (row one and two) and BraTS2024 (row three and row four) dataset for T2W images. The brighter areas represent region where the model is less confident.}
    \label{fig:comp}
\end{figure}

We compare UAMSA-UNet with baseline models, which serve as foundational architectures for medical image segmentation. To ensure a fair evaluation, we maintain a consistent experimental protocol across all datasets. The results, summarized in Table \ref{tab:full_comparison}, demonstrate that UAMSA-UNet outperforms other methods across all datasets. Fig. \ref{fig:comp} further illustrates this improvement on the T2W image modality, where our model more accurately predicts tumor and its boundaries in closer alignment with the ground truth.

\begin{figure}
    \centering
    \includegraphics[width=1\linewidth]{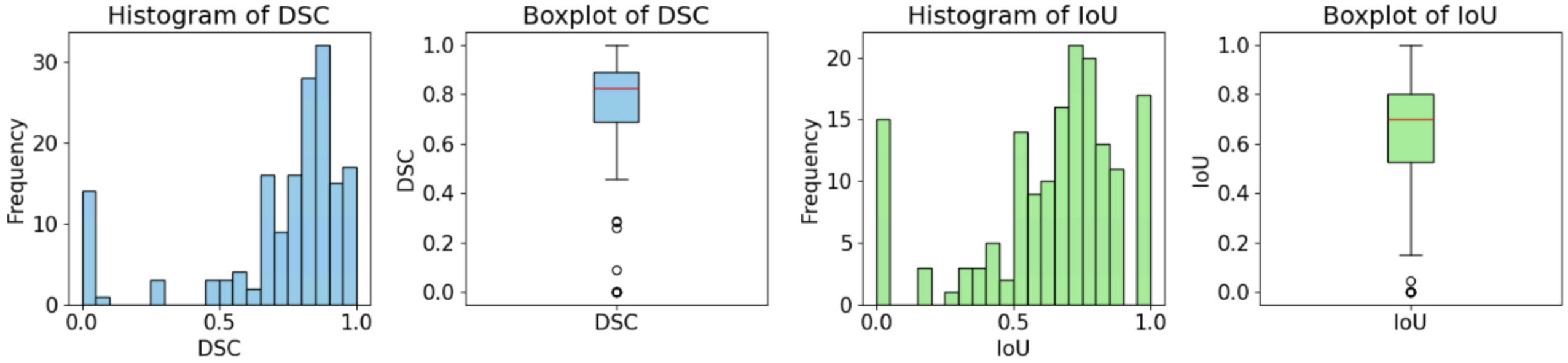}
    \caption{Histograms and boxplots of Dice (blue) and IoU (green) for T2W test images (BraTS2024). High central tendency indicates strong performance; outliers suggest rare segmentation errors.}
    \label{fig:hist_iou}
\end{figure}

Beyond segmentation performance, we assess computational efficiency by comparing hardware performance metrics and training resources across models, as detailed in Table \ref{tab:full_comparison}. The results indicate that UAMSA-UNet requires fewer FLOPs calculations, translating to higher throughput, as evident from the inference time across different image modalities.
Furthermore, the histograms and boxplots of Dice and IoU mterics in Fig. \ref{fig:hist_iou} reaffirm the model’s strong segmentation performance. Most Dice values cluster between 0.7 and 0.9, while IoU values fall predominantly in the 0.5–0.7 range, with median Dice ($\sim$0.80) and IoU ($\sim$0.70) reflecting consistent accuracy. Although a few outliers near zero indicate occasional segmentation failures, the overall results highlight the reliability and efficiency of UAMSA-UNet in brain tumor segmentation.

\section{Conclusion}
 UAMSA-UNet effectively enhances brain tumor segmentation by integrating multi-scale feature extraction, attention mechanisms, and Bayesian deep learning. It consistently outperforms traditional methods, delivering better accuracy and segmentation results, crucial for clinical decision-making. A key contribution, the uncertainty-aware loss function significantly improves segmentation performance as well as its reliability. One side-effect of our smoothing prior is that we may under-report true epistemic uncertainty in extremely ambiguous regions. In future work, we will explore hybrid losses that reward variance in boundary zones—e.g., by weighting the penalty term $\lambda_u$ according to edge confidence maps or by combining our learned smoothing with established calibration techniques such as temperature scaling.

 \section{Compliance with Ethical Standards}

This research study was conducted retrospectively using human subject data made available in open access. Ethical approval was not required as confirmed by the license attached with the open access data.

\bibliographystyle{IEEEbib}
\bibliography{strings,refs}

\end{document}